\documentclass[conference]{IEEEtran}
\IEEEoverridecommandlockouts
\usepackage[noadjust]{cite}
\usepackage{amsmath,amssymb,amsfonts}
\usepackage{algorithmic}
\usepackage{graphicx}
\usepackage{textcomp}
\usepackage{xcolor} 
\usepackage{multirow} 
\usepackage{array}
\usepackage{booktabs}
\usepackage{capt-of}
\usepackage{dblfloatfix}
\usepackage{xurl}
\usepackage{hyperref}

\hypersetup{
    colorlinks=true,
    urlcolor=linkblue,
    linkcolor=linkblue,
    citecolor=linkblue
}
\usepackage{xcolor}

\definecolor{linkblue}{RGB}{102,113,194}

\makeatletter
\let\UMR@makecaption\@makecaption
\long\def\@makecaption#1#2{%
  \ifx\@captype\@IEEEtablestring
    {\footnotesize\noindent #1: #2\par}%
    \@IEEEtablecaptionsepspace
  \else
    \UMR@makecaption{#1}{#2}%
  \fi}
\makeatother

\def\BibTeX{{\rm B\kern-.05em{\sc i\kern-.025em b}\kern-.08em
    T\kern-.1667em\lower.7ex\hbox{E}\kern-.125emX}}
\begin{document}

\title{\LARGE \bf
Unified Motion Retargeting for Humanoids with Learned Point Cloud Correspondence}

\author{
\IEEEauthorblockN{
\textbf{Hanyang Cao}\textsuperscript{1,2,*},
\textbf{Yuetong Fang}\textsuperscript{1,2,*},
\textbf{Taesoo Kwon}\textsuperscript{3,*},
\textbf{Runyi Yu}\textsuperscript{2,4},
\textbf{Ji Ma}\textsuperscript{5},
\textbf{Jing Tan}\textsuperscript{1,2},\\
\textbf{Yangchen Zhou}\textsuperscript{1},
\textbf{Baoze Du}\textsuperscript{2},
\textbf{Yi Gu}\textsuperscript{1},
\textbf{Yukang Gao}\textsuperscript{1,2},
\textbf{Ruoli Dai}\textsuperscript{2},
\textbf{Lei Han}\textsuperscript{2,\(\dagger\)},
\textbf{Renjing Xu}\textsuperscript{1,\(\dagger\)}
}

\IEEEauthorblockA{
\textsuperscript{1}HKUST(GZ) \quad
\textsuperscript{2}Noitom Robotics \quad
\textsuperscript{3}Hanyang University \quad
\textsuperscript{4}HKUST \quad
\textsuperscript{5}HKU\\[0.4em]
\textsuperscript{*}Equal contribution \quad
\textsuperscript{\(\dagger\)}Corresponding authors
}
}

\IEEEaftertitletext{%
    \vspace{-0.8cm}
    \begin{center}
        \href{https://hanyang9.github.io/UMR/}{%
            \textcolor{linkblue}{%
                \large\bfseries UMR Project Page
            }%
        }
    \end{center}
    % \vspace{0.1cm}
}

\maketitle

\thispagestyle{empty}
\pagestyle{empty}

\begin{abstract}
Humanoid learning increasingly relies on transforming vast and diverse human motion data into high-quality robot reference trajectories. 
However, retargeting human motion to humanoid robots is challenging due to substantial differences in morphology, degrees of freedom, joint ranges, and kinematic constraints between humans and robots.
Existing retargeting methods typically address these differences by defining human-robot correspondence through hand-crafted sparse keypoints or body-part pairs. As a result, retargeting quality depends heavily on manual semantic design, limiting scalability across motion sources and robot morphologies and providing only sparse guidance for reproducing detailed poses and interactions.
In this paper, we present Unified Motion Retargeting (UMR), a framework that learns dense point cloud correspondence without requiring manually designed human-robot mappings. By treating exterior point clouds as a unified interface between human motion and humanoid robots, UMR decouples retargeting from source-specific skeletal semantics and robot-specific topology. The learned dense correspondence provides fine-grained geometric anchors for constrained point cloud matching optimization, enabling surface-level pose alignment and direct transfer of interaction contacts. Experiments demonstrate that UMR unifies retargeting across heterogeneous motion sources, robot embodiments, and downstream scenarios ranging from locomotion to interaction, while achieving higher motion fidelity and plausibility than state-of-the-art methods. UMR therefore provides a scalable foundation for transforming large-scale human motion references into robot-ready training data.
\end{abstract}

\section{Introduction}

Humanoid robots are expected to perform the same kinds of whole-body behaviors that make human motion so expressive: walking, changing posture, manipulating objects, and making physical contact with the surrounding environment. Learning these behaviors requires large-scale demonstrations; however, directly collecting large amounts of high-quality motion data from robots is costly, time-consuming, and difficult to scale. Human motion provides a natural and scalable corpus of whole-body behavioral data, although directly using human motion as a reference trajectory for robots is generally infeasible due to the inherent differences in body proportions, joint topology, degrees of freedom, and kinematic constraints. 
As motion data grow in scale and diversity, retargeting becomes a critical interface between human motion corpora and humanoid behaviors, which must produce high-quality references while generalizing efficiently across motion representations and robot embodiments.

Existing humanoid motion-retargeting methods are predominantly skeleton-centric. Whether based on kinematic optimization~\cite{gmr}, learning-augmented adaptation~\cite{reactor}, or interaction-aware constraints~\cite{yang2025omniretarget}, these methods rely on a one-to-one mapping between manually selected human joints and the robot topologies. 
The resulting correspondences inherit the topology and semantics of the paired skeletons, making them specific to a particular embodiment. Switching to a new target robot therefore requires redefining the body-part mapping and readjusting the motion-fitting parameters. Moreover, sparse joint-level representations constrain motion at only a small subset of the body geometry, providing limited morphologic information for fine-grained pose alignment and reliable contact transfer.

\begin{figure}[tbp]
\centering
\includegraphics[width=0.95\linewidth, trim= 950 1150 1300 720, clip]{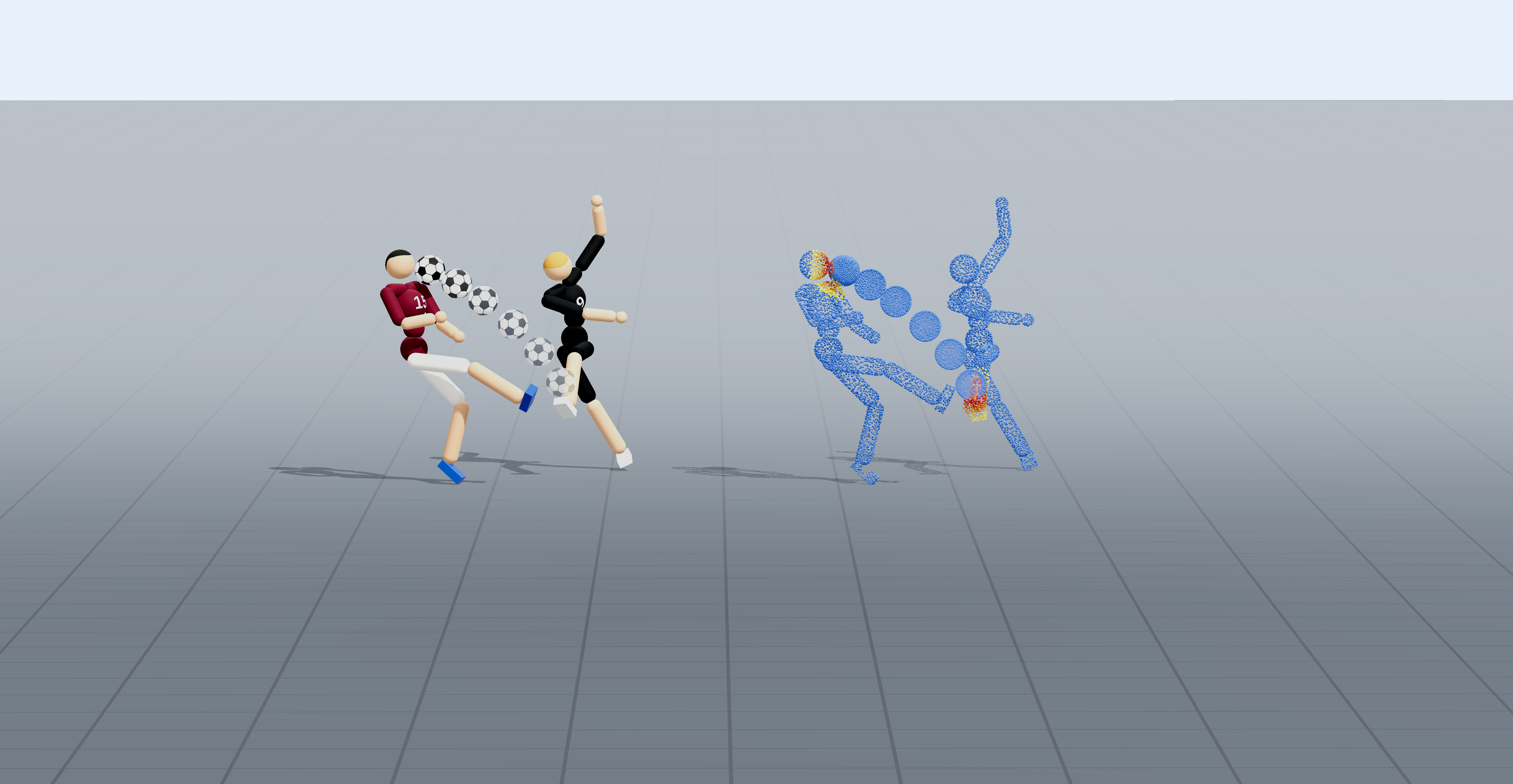}
\includegraphics[width=0.95\linewidth, trim= 800 1080 1400 820, clip]{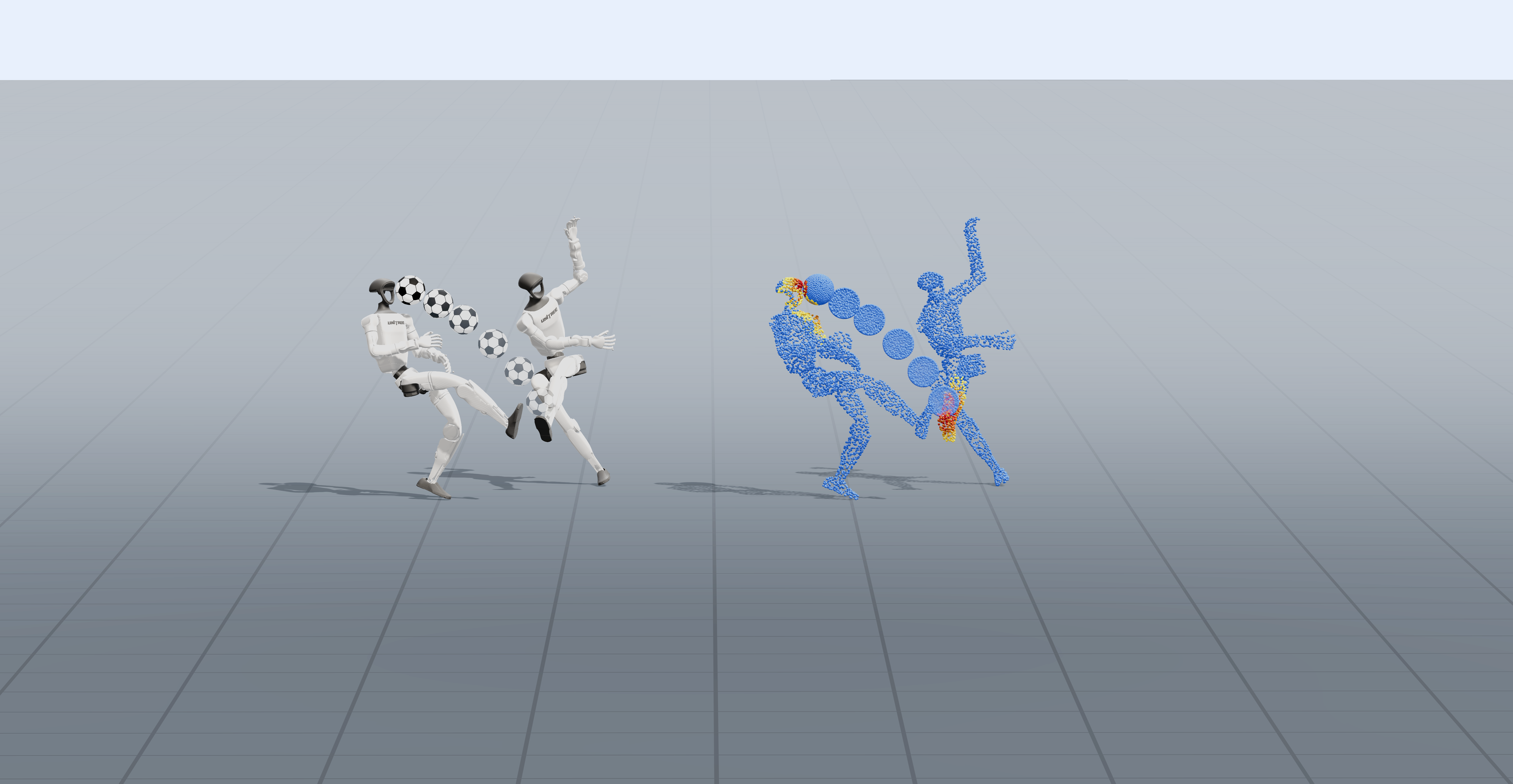}
\caption{\textbf{Visualization of motion retargeting with UMR}. Top: source human motion and its surface point clouds with contact maps. Bottom: the corresponding motion retargeted to the Unitree G1 humanoid and its surface point clouds with transferred contacts. By establishing dense surface correspondences, UMR transfers both motion and contact across substantially different embodiments without requiring manually specified joint correspondences. 
}
\label{fig:teaser}
\end{figure}

Meshes, in contrast,  directly represent the body through a densely sampled surface rather than isolated joints, allowing correspondence to be inferred from a shared spatial extent and local geometry, rather than being prescribed by specific joint semantics. 
Unlike skeletal mappings, which pair discrete joints, surface mappings distribute correspondence over the body using 3D positions and local orientations, including regions between joints. Surface correspondence can thus be established without requiring human and robot joints to match one-to-one. More importantly, meshes can directly expose contact-relevant geometry to the optimization. 
Recent human-motion representations increasingly represent motion through deformable human mesh models, such as SMPL-X~\cite{SMPL-X:2019}, MHR~\cite{MHR:2025}, and SOMA~\cite{saito2026soma}.

Turning a mesh motion into robot motion requires more than matching body shapes: each source surface region must be linked to the robot region that should follow it throughout the motion.
In this paper, we present Unified Motion Retargeting (UMR), a unified framework that couples learned point cloud correspondence with constrained motion optimization. UMR first learns an ordered correspondence between source and robot point clouds in a canonical pose and binds the paired points to their respective meshes. As the meshes move, the paired correspondence provide fine-grained targets for matching positions and surface orientations under kinematic constraints. Beyond pose matching, the same correspondence supports direct contact map transfer between embodiments, without additional hand-designed associations between human contact regions and robot body parts. As shown in Figure~\ref{fig:teaser}, this formulation provides a common retargeting interface across heterogeneous mesh-based motion sources and robot embodiments.

In summary, our contributions are as follows:
\begin{itemize} 
\item A unified surface-centric retargeting framework that learns dense source-robot correspondence from surface point clouds, avoiding manually defined skeletal or body mappings and supporting heterogeneous motion representations and robot embodiments. 

\item An efficient retargeting pipeline that uses the learned correspondence for surface-level pose matching and direct contact-map transfer, with a reusable correspondence setup shared across motions.

\item A systematic downstream evaluation of retargeted reference quality across whole-body tracking, large-scale policy learning, and contact-rich interaction tasks.

\end{itemize}

\begin{figure*}[tbp]
\centering
\includegraphics[width=1.0\linewidth, trim=0 0 0 0, clip]{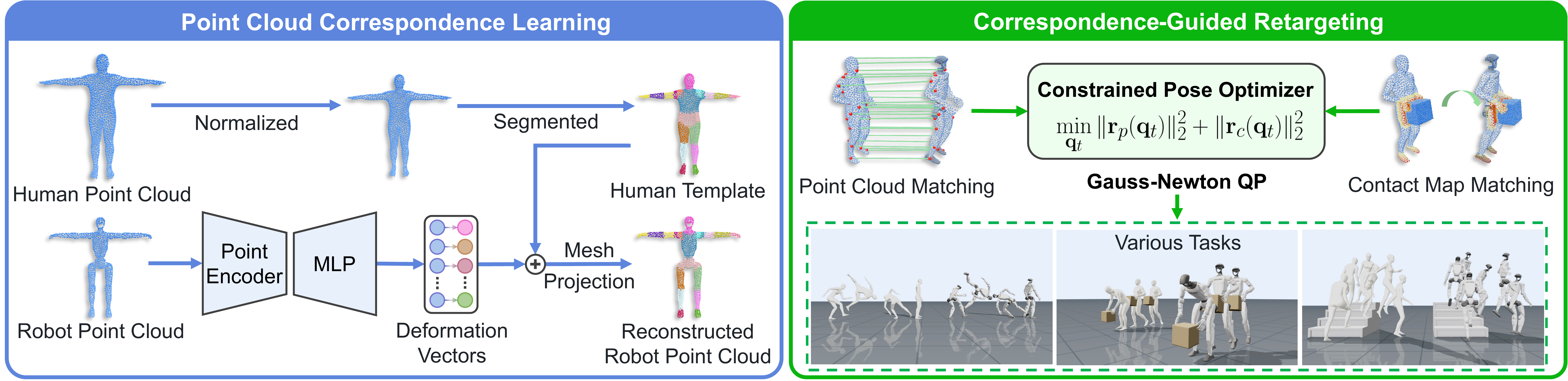}
\caption{\textbf{UMR overview.} UMR establishes a shared surface-based interface for heterogeneous motion sources and robot embodiments through two stages. Point Cloud Correspondence Learning (left) learns a reusable indexed correspondence between canonical source and robot point clouds and binds the paired points to their body meshes.
Correspondence-Guided Retargeting (right) reuses the same point pairs to match surface positions, orientations, and contact maps in a constrained pose optimizer, producing robot motions for diverse tasks covering locomotion and interaction.
}

\label{fig:overview}
\end{figure*}

\section{Related Work}
\subsection{Human-to-humanoid Motion Retargeting}

Motion retargeting was initially studied in computer animation as a way to transfer motion-capture data between articulated characters with different segment lengths and proportions. Early methods treated this transfer as a sequence-level spacetime optimization problem that preserved selected motion features across body proportions~\cite{10.1145/280814.280820}. As retargeting moved toward interactive use, subsequent methods recast the same feature-preservation problem as online tracking, using inverse-rate control and sequential filtering to follow selected end-effector trajectories under kinematic and dynamic constraints \cite{choi2000online,tak2005physically}. These studies established the optimization-based foundation of motion retargeting, with correspondence typically defined in a skeleton-centric manner through selected joints or end effectors.

Humanoid-robot retargeting builds on this approach by incorporating the kinematic, dynamic, and physical constraints required for robot control. Within this setting, prior work has pursued complementary routes: enforcing balance and whole-body feasibility~\cite{penco2018robust,darvish2019whole}, adapting the target morphology jointly with the motion~\cite{ayusawa2017motion}, and reducing retargeting artifacts through non-uniform scaling and staged optimization~\cite{gmr}.
Interaction-aware methods further introduce kinematic and contact constraints for object and scene interactions \cite{yang2025omniretarget}. Learning-based methods fit human-robot joint mappings from paired or unpaired data, often through shared latent representations~\cite{yan2023imitationnet,reactor, chen2026scalablewholebodymotiontransfer}. 

Although these methods differ in their objectives, solvers, and adaptation mechanisms, cross-embodiment correspondence remains predominantly skeleton-centric, defined over selected joints, links, keypoints, or rigid-body mappings. Supporting a new embodiment therefore commonly requires redefining
the skeletal correspondence and retuning embodiment-specific fitting objectives. To avoid this embodiment-specific re-matching,  UMR instead learns indexed correspondence between exterior body point clouds and uses the resulting point pairs as motion-matching anchors, thus removing the need for predefined skeletal keypoints or rigid body mappings.

\subsection{Point Cloud-Based Body Correspondence}

Point cloud correspondence is widely used to register observed human geometry to a canonical body model. These associations turn observed surface geometry into constraints for estimating the pose and shape of the model. Existing methods differ primarily in how the associations are established. Template-deformation methods warp a canonical surface toward the observation~\cite{groueix20183d}, while implicit-field methods map observed points into canonical body coordinates~\cite{bhatnagar2020loopreg,wang2021locally}. Other approaches iteratively refine the body template against the observation~\cite{corona2022learned,marin2024nicp}, or predict surface markers and landmarks as intermediate fitting targets~\cite{li2025etch,cai2026omnifit}. When applied to body fitting, these methods operate with a predefined human template and compatible anatomy. Human-to-robot retargeting instead relates bodies with different surface geometry and kinematic structures, requiring correspondence that can be established without a shared body model and remain consistent throughout motion.

Existing work has developed two components of surface-based retargeting separately. One line incorporates geometry into motion transfer while retaining a skeletal or marker-based structure. Geometry-aware deformation is combined with skeleton-aware transfer~\cite{rekik2024correspondence}, contact-rich hand retargeting relies on surface matching and sparse markers \cite{lakshmipathy2025kinematic}, and MeshRet derives dense surface sensors from sparse skeletal correspondences~\cite{ye2024skinned}. A second line represents contact over sampled surfaces: HDM uses coherently ordered object point clouds for contact transfer~\cite{xie2023template}, while BimArt uses surface-based contact maps for interaction synthesis~\cite{zhang2025bimart}. What remains missing is a whole-body surface mapping across embodiments. UMR fills this gap by learning such correspondence without a predefined skeletal map and using the resulting point pairs for constrained motion retargeting and direct contact map transfer.

\section{Method}
\subsection{Overview}

UMR represents motion through the moving exterior body surface rather than the source skeleton. Any motion source that provides a canonical T-pose mesh and its posed surfaces over time can therefore enter the same pipeline, including SMPL-family models~\cite{SMPL-X:2019}, SOMA representations~\cite{saito2026soma}, rigged humanoid characters~\cite{peng2018deepmimic}, and scanned human meshes. Sampling these surfaces as point clouds avoids requiring compatible skeletons or mesh topology between the source and robot, while retaining the geometry needed to describe pose and contact. Given the resulting source sequence and a target humanoid model, UMR estimates robot generalized coordinates whose surface motion follows the source while satisfying contact and kinematic constraints.

Turning surface motion into robot motion requires two steps: determining which surface locations correspond across embodiments and using those correspondences to compute feasible robot poses. The UMR pipeline mirrors these requirements. As illustrated in Figure~\ref{fig:overview}, UMR decomposes motion retargeting into two corresponding stages. Point Cloud Correspondence Learning establishes an ordered mapping between aligned source and robot point clouds in their canonical T-poses and binds each point pair to the corresponding body meshes, allowing the mapping to persist as both bodies move. Then, Correspondence-Guided Retargeting optimizes the robot motion frame-by-frame by matching the positions, surface orientations, and contacts of the paired points while maintaining temporal consistency and respecting kinematic constraints. Using the same surface correspondence throughout provides a common retargeting formulation across motion sources and robot embodiments.

\begin{figure*}[tbp]
\centering
\includegraphics[width=1.0\linewidth, trim= 0 0 0 0, clip]{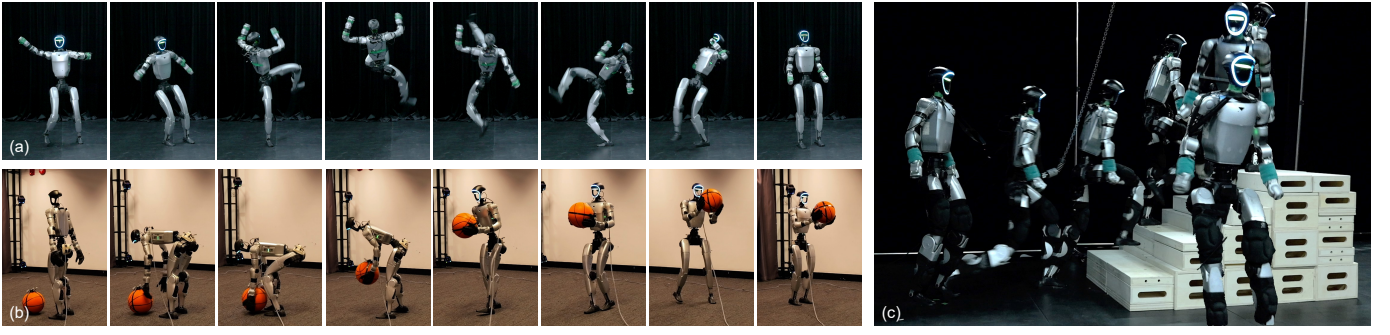}
\caption{\textbf{Real world Deployment.} High-quality references generated by UMR support successful policy training and real-world deployment across high-dynamic motions and diverse interaction scenarios, demonstrating reliable transfer from retargeted motion data to physical humanoid control. (a) Spin kick. (b) Ball pick-up followed by backward walking with turns, deployed with a motion capture system. (c) Stair traversal followed by jumping down.}
\label{fig:2real}
\end{figure*}

\subsection{Point Cloud Correspondence Learning}
\label{sec:correspondence-learning}
Point Cloud Correspondence Learning aims to produce indexed human-robot surface pairs that can be reused throughout a motion sequence. UMR learns this mapping once from aligned canonical T-poses: the paired points are later attached to their respective meshes and transported with the bodies as they move. Learning in the canonical pose isolates stable geometric differences between the two embodiments from pose-dependent deformation.

For a given source representation, a single human template defines the source point indices shared across all motion frames. Different source formats enter through the same interface consisting of a canonical template and its posed point cloud sequence. Let $\mathbf X^h=\{\mathbf x^h_i\in\mathbb R^3\}_{i=1}^N$ denote the ordered exterior point cloud sampled from the aligned human T-pose, and let $\mathbf X^r=\{\mathbf x^r_j\in\mathbb R^3\}_{j=1}^N$ denote an unordered exterior point cloud sampled from the aligned robot T-pose. The ordering of $\mathbf X^h$ provides the indices that define the learned correspondence, whereas $\mathbf X^r$ supplies the target surface geometry. A PointNet-style encoder $E_\theta$ summarizes $\mathbf X^r$ into a latent representation~\cite{qi2017pointnet}, and an MLP decoder $D_\theta$ predicts one deformation vector for each indexed human point, producing its corresponding location on the robot:
\begin{equation}
\hat{\mathbf X}^r = \mathbf X^h + D_\theta(E_\theta(\mathbf X^r)).
\end{equation}
where $\hat{\mathbf X}^r=\{\hat{\mathbf x}^r_i\}_{i=1}^N$ is the reconstructed robot-side correspondence point cloud, $\mathbf d_i$ is the deformation vector applied to the $i$-th human point, and $\hat{\mathbf x}^r_i=\mathbf x^h_i+\mathbf d_i$. The index $i$ is therefore inherited from the human point cloud and defines a correspondence point shared by the human point cloud and the target robot point cloud.

\textbf{Training Objective.} The correspondence loss $\mathcal L_{\mathrm{corr}}$ encourages the reconstructed point cloud to cover the target point cloud while maintaining locally smooth deformation vectors over the human-template geodesic graph:
\begin{equation}
\mathcal L_{\mathrm{corr}} =
\lambda_c \mathcal L_c
+ \lambda_r \mathcal L_r
+ \lambda_e \mathcal L_e .
\end{equation}

The Chamfer term $\mathcal L_c$ provides the primary geometric supervision by minimizing the symmetric squared nearest-neighbor distance between the reconstructed point cloud $\hat{\mathbf X}^{r}$ and the unordered target robot point cloud
$\mathbf X^r$:
\begin{equation}
\mathcal L_c =
\frac{1}{N}\sum_{i=1}^N \min_{1\le j\le N}
\|\hat{\mathbf x}^r_i-\mathbf x^r_j\|_2^2
+ \frac{1}{N}\sum_{j=1}^N \min_{1\le i\le N}
\|\mathbf x^r_j-\hat{\mathbf x}^r_i\|_2^2 .
\end{equation}

The repulsion term $\mathcal L_r$ discourages multiple points from collapsing to the same local point cloud region:
\begin{equation}
\mathcal L_r =
\frac{1}{N K_r}\sum_{i=1}^N
\sum_{\ell\in\mathcal N_r(i)}
\exp(-\|\hat{\mathbf x}^r_i-\hat{\mathbf x}^r_\ell\|_2^2/r^2),
\end{equation}
where $\mathcal N_r(i)$ denotes the $K_r$ nearest reconstructed points to point $i$.

The edge smoothness term $\mathcal L_e$ is essential for learning coherent point cloud correspondence by enforcing similar deformation vectors between human-template points connected by the edge set $\mathcal E$ of a fixed geodesic graph:
\begin{equation}
\mathcal L_e =
\frac{1}{|\mathcal E|}\sum_{(i,\ell)\in \mathcal E}
\|\mathbf d_i-\mathbf d_\ell\|_2^2.
\end{equation}

As a result, the learned human and robot point clouds retain a shared ordering, allowing robot points to inherit human body segment labels. This representation provides a unified interface for the subsequent optimizer, where segment-specific weights and parameters defined on the human template can be reused across different robots without additional body mappings.

\subsection{Correspondence-Guided Retargeting}
\label{sec:correspondence-retargeting}
The canonical point pairs become motion targets after being bound to their respective T-pose meshes. During motion, human points follow the posed source mesh through barycentric transport, while robot points follow their link-local bindings through forward kinematics. Each source point therefore supplies a motion target for its robot counterpart, turning retargeting into a surface matching problem.

\textbf{Optimization Objective.} 
Differences in geometry and kinematics prevent the robot from matching every surface target exactly. At each frame, UMR balances whole-body pose matching with additional emphasis on active contacts, and estimates robot generalized coordinates $\mathbf q_t$ by minimizing a stack of point-level residuals:
\begin{equation}
  \min_{\mathbf q_t}
  \|\mathbf r_p(\mathbf q_t)\|_2^2
  +\|\mathbf r_c(\mathbf q_t)\|_2^2 .
\end{equation}
The pose residual $\mathbf r_{p,i}$ matches both position and local surface orientation:
\begin{equation}
  \mathbf r_{p,i}(\mathbf q_t)=
  \begin{bmatrix}
  \sqrt{w_i^p}\bigl(\mathbf x^r_i(\mathbf q_t)-\mathbf x^h_{t,i}\bigr)\\
  \sqrt{w_i^n}\bigl(\bar{\mathbf n}^r_i(\mathbf q_t)
  -\bar{\mathbf n}^h_{t,i}\bigr)
  \end{bmatrix},
  \qquad i\in\mathcal I,
\end{equation}
where $\mathcal I$ denotes the selected set of correspondence points. The normal offsets $\bar{\mathbf n}^h_{t,i}$ and $\bar{\mathbf n}^r_i(\mathbf q_t)$ measure orientation changes relative to their respective T-pose bindings. The segment-dependent weights $w_i^p$ and $w_i^n$ balance the contributions of different human body regions.

\begin{figure}[tbp]
\centering
\includegraphics[width=0.6\linewidth, trim= 0 0 100 0, clip]{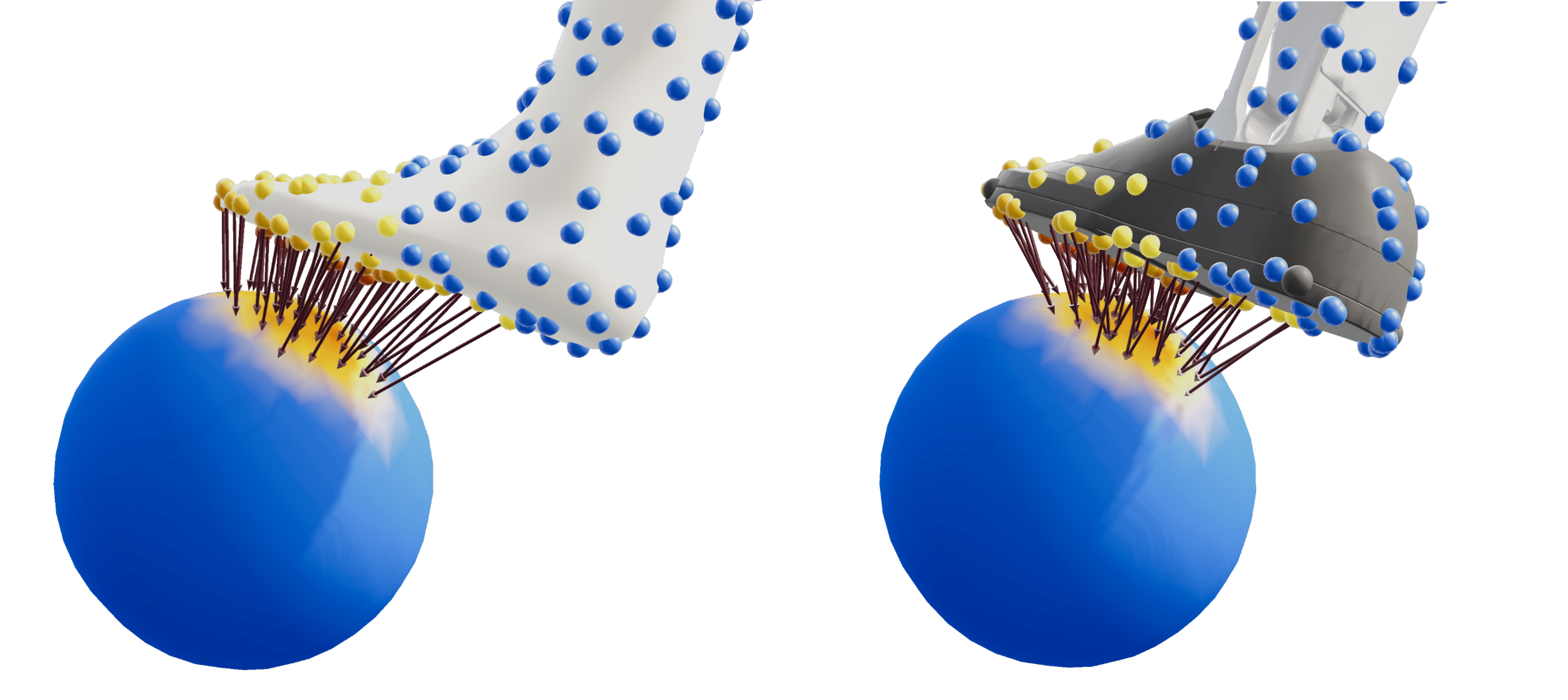}
\caption{\textbf{Visualization of contact map.} Contact maps are represented by directional vectors from each keypoint in the point cloud to its nearest point on the object surface. Surface correspondence enables direct transfer of these contact relationships from the source motion (left) to the target robot (right).}
\label{fig:vis_contact_map}
\end{figure}

\begin{figure*}[tbp]
\centering
\includegraphics[width=1.0\linewidth, trim= 300 670 0 630, clip]{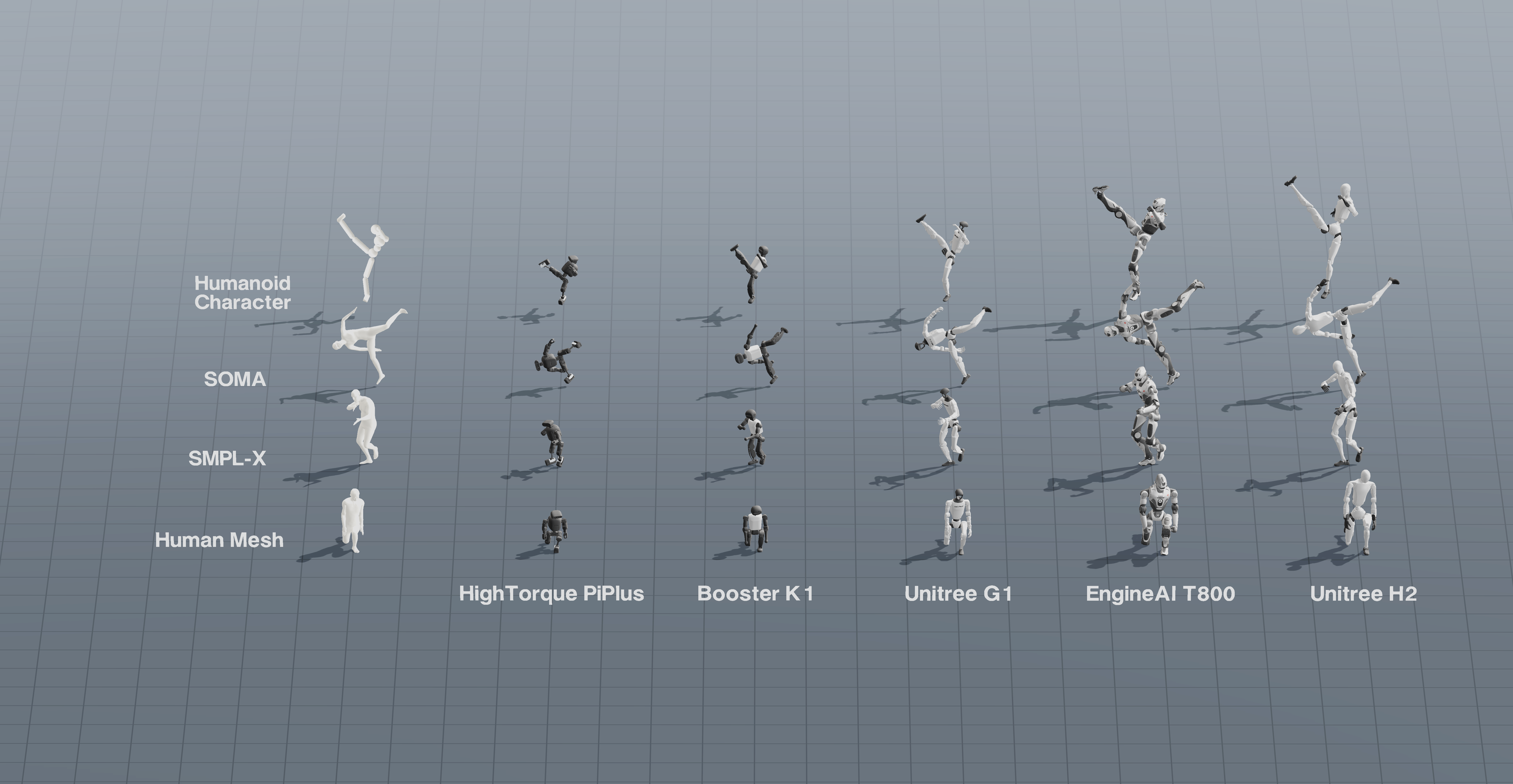}
\caption{\textbf{Unified retargeting across motion sources and robot embodiments.} Qualitative results across four source representations and five humanoid embodiments demonstrate that UMR preserves high motion fidelity across heterogeneous source geometries and robot kinematic configurations.}
\label{fig:unified-sources-embodiments}
\end{figure*}

The contact residual $\mathbf r_{c,i}$ matches a contact map between the human and robot point clouds using the same point indexing without an additional body map:
\begin{equation}
\mathbf r_{c,i}(\mathbf q_t)=
\sqrt{w_i^c}\left(
\mathbf c^r_i(\mathbf q_t)-\mathbf c^h_{t,i}
\right),
\qquad i\in\mathcal C_t .
\end{equation}
Following BimArt~\cite{zhang2025bimart}, we represent the contact map using paired human and robot contact vectors, as illustrated in Fig.~\ref{fig:vis_contact_map}. Let $\mathbf Y_t=\{\mathbf y_{t,j}\in\mathbb R^3\}_{j=1}^M$ denote the environment point cloud sampled from an interacting object or scene surface. For each human correspondence point $i\in\mathcal I$, its nearest environment point is:
\begin{equation}
\begin{aligned}
\pi_t(i)
&=\arg\min_{1\leq j\leq M}
\|\mathbf x^h_{t,i}-\mathbf y_{t,j}\|_2.
\end{aligned}
\end{equation}
Using this shared environment-point assignment, the contact map is defined as:
\begin{equation}
\begin{aligned}
\mathbf c^h_{t,i}
&=\mathbf x^h_{t,i}-\mathbf y_{t,\pi_t(i)},\\
\mathbf c^r_i(\mathbf q_t)
&=\mathbf x^r_i(\mathbf q_t)-\mathbf y_{t,\pi_t(i)}.
\end{aligned}
\end{equation}
The active contact map set contains correspondence points within the contact threshold $\tau_c$:
\begin{equation}
\mathcal C_t
=
\left\{
i\in\mathcal I
\;\middle|\;
\left\|\mathbf c^h_{t,i}\right\|_2\leq\tau_c
\right\}.
\end{equation}
The same construction applies uniformly to contacts with the ground, manipulated objects, and surrounding scene geometry. For self-contact, the external environment point is replaced by another correspondence point located on a distinct, non-adjacent body segment.

\textbf{Constrained Update.}
The point residuals depend nonlinearly on $\mathbf q_t$ through forward
kinematics. We address these requirements through a constrained Gauss-Newton Quadratic Programming.

At each iteration, the pose residuals $\mathbf r_p(\mathbf q_t)$ and contact residuals $\mathbf r_c(\mathbf q_t)$ are stacked into $\mathbf r(\mathbf q_t)$ and linearized as:
\begin{equation}
\begin{aligned}
  \mathbf r(\mathbf q_t+\Delta\mathbf q)
  &\approx
  \mathbf r(\mathbf q_t)+\mathbf J(\mathbf q_t)\Delta\mathbf q,\\
  \mathbf J(\mathbf q_t)
  &=
  \left.
  \frac{\partial\mathbf r(\mathbf q)}
  {\partial\mathbf q}
  \right|_{\mathbf q=\mathbf q_t}.
\end{aligned}
\end{equation}
The resulting update $\Delta\mathbf q$ is a damped constrained Gauss-Newton step, formulated as a convex quadratic subproblem:
\begin{equation}
\begin{aligned}
\min_{\Delta\mathbf q} \quad
& \frac{1}{2}\|\mathbf r(\mathbf q_t)+\mathbf J(\mathbf q_t)\Delta\mathbf q\|_2^2+\frac{\mu}{2}\|\Delta\mathbf q\|_2^2 \\
\text{s.t.}\quad
& \mathbf q^-\le \mathbf q_t+\Delta\mathbf q\le \mathbf q^+,\\
& \mathbf A_t\Delta\mathbf q\le \mathbf b_t,\\
& \|\Delta\mathbf q\|_2 \le \eta .
\end{aligned}
\label{eq:constrained_update}
\end{equation}
Here, $\mu>0$ denotes the damping coefficient that regularizes the local update. $\mathbf q^-$ and $\mathbf q^+$ represent the lower and upper joint limits, respectively. $\eta$ refers to the trust-region radius. The linear inequality
$\mathbf A_t\Delta\mathbf q\leq\mathbf b_t$
enforces floor clearance.

Specifically, let $\mathcal F_t$ denote the active robot surface points
near the floor height $z_f$, and let $\mathbf J_i^z(\mathbf q_t)$ be
the Jacobian of their height $z_i^r(\mathbf q_t)$. Linearizing
$z_i^r(\mathbf q_t+\Delta\mathbf q)\geq z_f$ gives:
\begin{equation}
  -\mathbf J_i^z(\mathbf q_t)\Delta\mathbf q
  \leq
  z_i^r(\mathbf q_t)-z_f,
  \qquad i\in\mathcal F_t.
\end{equation}
Stacking these inequalities yields $\mathbf A_t\Delta\mathbf q\leq\mathbf b_t$ in Eq.~\ref{eq:constrained_update}.

\section{Experiments}

In this section, we evaluate UMR from three complementary perspectives. We first demonstrate its unified retargeting capability across heterogeneous motion representations and robot embodiments. We then examine the quality of the resulting references from per-motion tracking to large-scale policy learning, and further evaluate its effectiveness in contact-rich human-object and human-scene interactions. All quantitative evaluations are conducted on the Unitree G1.

UMR is implemented with MuJoCo~\cite{todorov2012mujoco}, with the constrained Gauss-Newton subproblems solved using Clarabel~\cite{goulart2026clarabel}. UMR also enables efficient motion retargeting with a reusable correspondence setup. Detailed runtime statistics, measured on an NVIDIA GeForce RTX~4070 Ti SUPER GPU and an Intel Core Ultra 7 265KF CPU, are reported in Table~\ref{tab:umr-efficiency}. All downstream RL policies are trained on NVIDIA RTX~4090 GPUs.

\begin{table}[t]
    \caption{Computational efficiency of UMR averaged over the LAFAN1 dataset~\cite{harvey2020robust}. }
    \label{tab:umr-efficiency}
    \centering
    \scriptsize
    \setlength{\tabcolsep}{2pt}
    \setlength{\arrayrulewidth}{0.4pt}
    \setlength{\heavyrulewidth}{0.6pt}
    \setlength{\aboverulesep}{0.4ex}
    \setlength{\belowrulesep}{0.4ex}
    \renewcommand{\arraystretch}{1.08}
    \begin{tabular}{@{}
    >{\hspace*{4pt}\centering\arraybackslash}m{0.15\columnwidth}
    >{\raggedright\arraybackslash}m{0.47\columnwidth}
    >{\centering\arraybackslash}m{0.30\columnwidth}
    @{}}
    \toprule
    \textbf{Stage}
    & \textbf{System Component}
    & \textbf{Computational Cost} \\
    \specialrule{\arrayrulewidth}{0.4ex}{0.4ex}

    \multirow{5}{*}{\textbf{Stage I}}
    & \multicolumn{2}{
        >{\raggedright\arraybackslash}
        m{\dimexpr0.77\columnwidth+2\tabcolsep\relax}
      }{\hspace*{0pt}\textit{Reusable Point Cloud Correspondence Setup}} \\
    \cmidrule(l{2pt}r{0pt}){2-3}
    & Point cloud sampling          & 9.83 s  \\
    & Geodesic precomputation   & 5.58 s  \\
    & Correspondence training   & 10.38 s \\
    \cmidrule(l{2pt}r{0pt}){2-3}
    & \textbf{Total setup time} & 25.79 s \\
    \specialrule{\arrayrulewidth}{0.4ex}{0.4ex}

    \multirow{4}{*}{\textbf{Stage II}}
    & \multicolumn{2}{
        >{\raggedright\arraybackslash}
        m{\dimexpr0.77\columnwidth+2\tabcolsep\relax}
      }{\hspace*{0pt}\textit{Per-clip Motion Retargeting}} \\
    \cmidrule(l{2pt}r{0pt}){2-3}
    & Data preprocessing               & 141.46 FPS \\
    & Motion retargeting               & 121.26 FPS \\
    \cmidrule(l{2pt}r{0pt}){2-3}
    & \textbf{Overall retargeting throughput} & 65.29 FPS \\
    \bottomrule
    \end{tabular}
\end{table}

\begin{table*}[t]
\caption{Motion tracking success rates (\%) on the LAFAN1 dataset~\cite{harvey2020robust}. Each entry is computed from 4,096 trials. A trial is successful if the policy completes the fixed reference window without triggering the termination criterion. The best values are \textbf{bold} and the second best are \underline{underlined}.}
  \label{tab:beyondmimic-success-rates}
  \centering
  \setlength{\tabcolsep}{4pt}
  \setlength{\arrayrulewidth}{0.4pt}
\setlength{\extrarowheight}{0pt}
\fontsize{4pt}{5pt}\selectfont
     \renewcommand{\arraystretch}{1.05}
  \resizebox{\textwidth}{!}{%
    \begin{tabular}{cccc|ccc|ccc}
      \hline
      \multirow{2}{*}{Motion}
        & \multicolumn{3}{c|}{Sim (w/o DR)}
        & \multicolumn{3}{c|}{Sim (w/ DR)}
        & \multicolumn{3}{c}{Sim2Sim} \\
      \cline{2-10}
        & UMR & GMR & Unitree
        & UMR & GMR & Unitree
        & UMR & GMR & Unitree \\
      \hline
      Dance
        & \underline{99.353}
        & 97.546
        & \textbf{99.396}
        & \textbf{98.941}
        & 95.065
        & \underline{98.682}
        & \textbf{96.793}
        & 89.682
        & \underline{96.100} \\
      Fall and GetUp
        & \underline{96.981}
        & 84.717
        & \textbf{97.046}
        & \underline{95.618}
        & 81.604
        & \textbf{95.667}
        & \underline{34.920}
        & 32.096
        & \textbf{46.899} \\
      Fight
        & \textbf{99.941}
        & 87.603
        & \underline{99.863}
        & \textbf{98.862}
        & 83.042
        & \underline{98.750}
        & \underline{95.298}
        & 79.492
        & \textbf{95.913} \\
      Jump
        & 99.577
        & \underline{99.618}
        & \textbf{99.992}
        & \underline{99.146}
        & 98.893
        & \textbf{99.512}
        & \textbf{92.863}
        & 89.266
        & \underline{91.528} \\
      Run
        & \textbf{99.988}
        & \underline{99.951}
        & \textbf{99.988}
        & \textbf{98.572}
        & 98.010
        & \underline{98.560}
        & \textbf{91.187}
        & 81.244
        & \underline{89.398} \\
      Sprint
        & \textbf{100.000}
        & 99.939
        & \underline{99.976}
        & \textbf{99.097}
        & 97.717
        & \underline{98.218}
        & \underline{86.426}
        & 71.460
        & \textbf{90.833} \\
      Walk
        & 98.977
        & \underline{98.993}
        & \textbf{99.988}
        & \underline{98.706}
        & 98.612
        & \textbf{99.770}
        & \underline{92.759}
        & 86.839
        & \textbf{95.188} \\
      \hline
    \end{tabular}%
  }
\end{table*}

\begin{table*}[t]
\caption{Motion tracking errors on the LAFAN1 dataset. Each cell reports the statistic over 4,096 trials as \textbf{Sim (w/o DR) / Sim (w/ DR)}. }
\label{tab:beyondmimic-tracking-errors}
\centering
\setlength{\tabcolsep}{4pt}
\setlength{\arrayrulewidth}{0.6pt}
\renewcommand{\arraystretch}{1.15}
\resizebox{\textwidth}{!}{%
\begin{tabular}{cccc|ccc|ccc}
\hline
\multirow{2}{*}{Statistics}
& \multicolumn{3}{c|}{$E_{\mathrm{g\text{-}mpbpe}}\,(\mathrm{mm})~\downarrow$}
& \multicolumn{3}{c|}{$E_{\mathrm{mpbpe}}\,(\mathrm{mm})~\downarrow$}
& \multicolumn{3}{c}{$E_{\mathrm{mpjpe}}\,(10^{-3}\,\mathrm{rad})~\downarrow$} \\
\cline{2-10}
& UMR & GMR & Unitree
& UMR & GMR & Unitree
& UMR & GMR & Unitree \\
\hline
Min
& \underline{44.94}/\underline{97.72}
& 46.44/106.76
& \textbf{34.39}/\textbf{95.38}
& \underline{16.73}/\underline{21.85}
& 18.86/23.99
& \textbf{15.47}/\textbf{21.05}
& \textbf{391.16}/\textbf{426.10}
& 414.45/459.75
& \underline{412.36}/\underline{452.75} \\
Median
& \underline{77.79}/\textbf{166.47}
& 121.08/230.95
& \textbf{77.01}/\underline{171.38}
& \underline{28.27}/\underline{35.05}
& 31.90/40.47
& \textbf{27.03}/\textbf{34.51}
& \textbf{581.68}/\textbf{627.80}
& 681.13/738.72
& \underline{609.38}/\underline{643.55} \\
Mean
& \underline{89.61}/\textbf{176.79}
& 198.89/322.57
& \textbf{83.99}/\underline{177.52}
& \underline{29.71}/\underline{36.81}
& 43.23/53.36
& \textbf{28.63}/\textbf{35.62}
& \textbf{610.19}/\textbf{648.77}
& 758.52/808.62
& \underline{620.05}/\underline{658.60} \\
Max
& \underline{236.49}/\underline{350.57}
& 843.48/989.32
& \textbf{171.85}/\textbf{265.66}
& \underline{53.42}/\underline{64.89}
& 150.70/166.97
& \textbf{53.34}/\textbf{61.65}
& \underline{984.03}/\underline{1031.10}
& 1321.24/1389.29
& \textbf{929.33}/\textbf{943.88} \\
\hline
\end{tabular}%
}
\end{table*}

\begin{figure}[tbp]
\centering
\includegraphics[width=0.85\linewidth, trim= 0 0 0 0, clip]{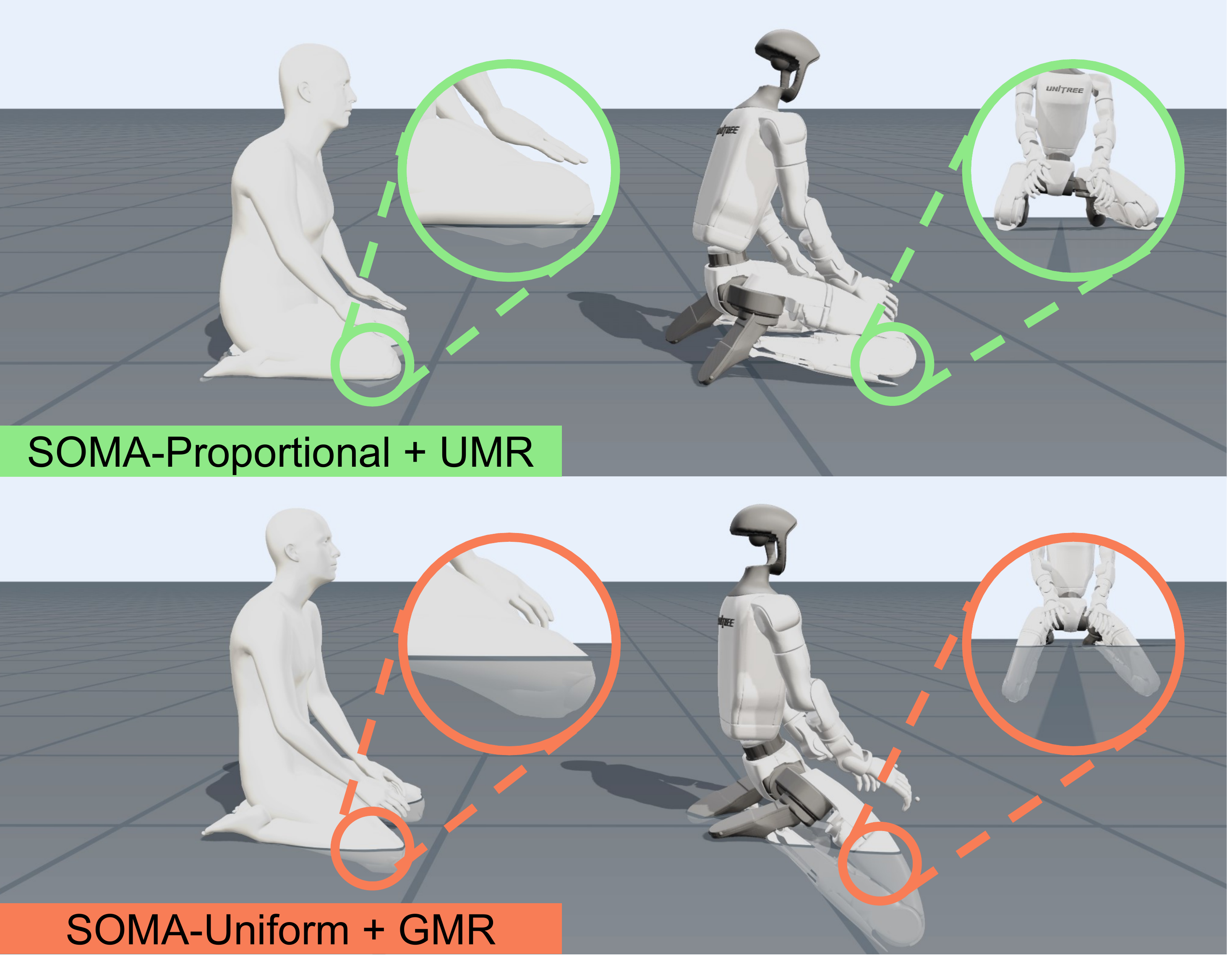}
\caption{\textbf{BONES-SEED retargeting comparison.} SOMA Proportional reference motion retargeted by UMR (top) and SOMA Uniform reference motion retargeted by GMR (bottom).}
\label{fig:saito2026soma-comparison}
\end{figure}

\subsection{Unified Retargeting Across Sources and Embodiments}

UMR supports motion sources with diverse underlying representations through a unified surface-based interface. Figure~\ref{fig:unified-sources-embodiments} illustrates this shared interface by showing retargeted motions from four source representations on five humanoid embodiments. The four sources comprise humanoid-character animations from MimicKit~\cite{MimicKitPeng2025}, SOMA motions from BONES-SEED~\cite{studio2026bones,saito2026soma}, SMPL-X motions from LaFAN1~\cite{harvey2020robust}, and motion sequences from a human mesh that we scanned and animated using in-house motion-capture data.
The selected embodiments span a broad range of body morphologies, with robot heights varying from 0.75~m to 1.83~m. Correspondence learning and retargeting remain unchanged across all source-robot pairs, with source-specific differences limited to the availability and definition of body segments used for point selection and weighting. SMPL-X and SOMA share the same body-region definition, the humanoid character uses partitions from its simplified geometry, and the unsegmented human mesh is treated as a single whole-body region.

\subsection{Whole-Body Tracking Performance Comparison}

We evaluate UMR against state-of-the-art baselines at two scales. First, we compare UMR with GMR~\cite{gmr} and Unitree retargeted references~\cite{unitree_lafan1_retargeting_2025} in per-motion tracking on LAFAN1 using BeyondMimic~\cite{liao2025beyondmimicmotiontrackingversatile}. Second, we evaluate the utility of UMR-retargeted data for large-scale policy learning on BONES-SEED under the SONIC framework~\cite{luo2025sonic}.

\begin{figure*}[htbp]
\centering
\includegraphics[width=0.32\linewidth, trim= 0 0 0 0, clip]{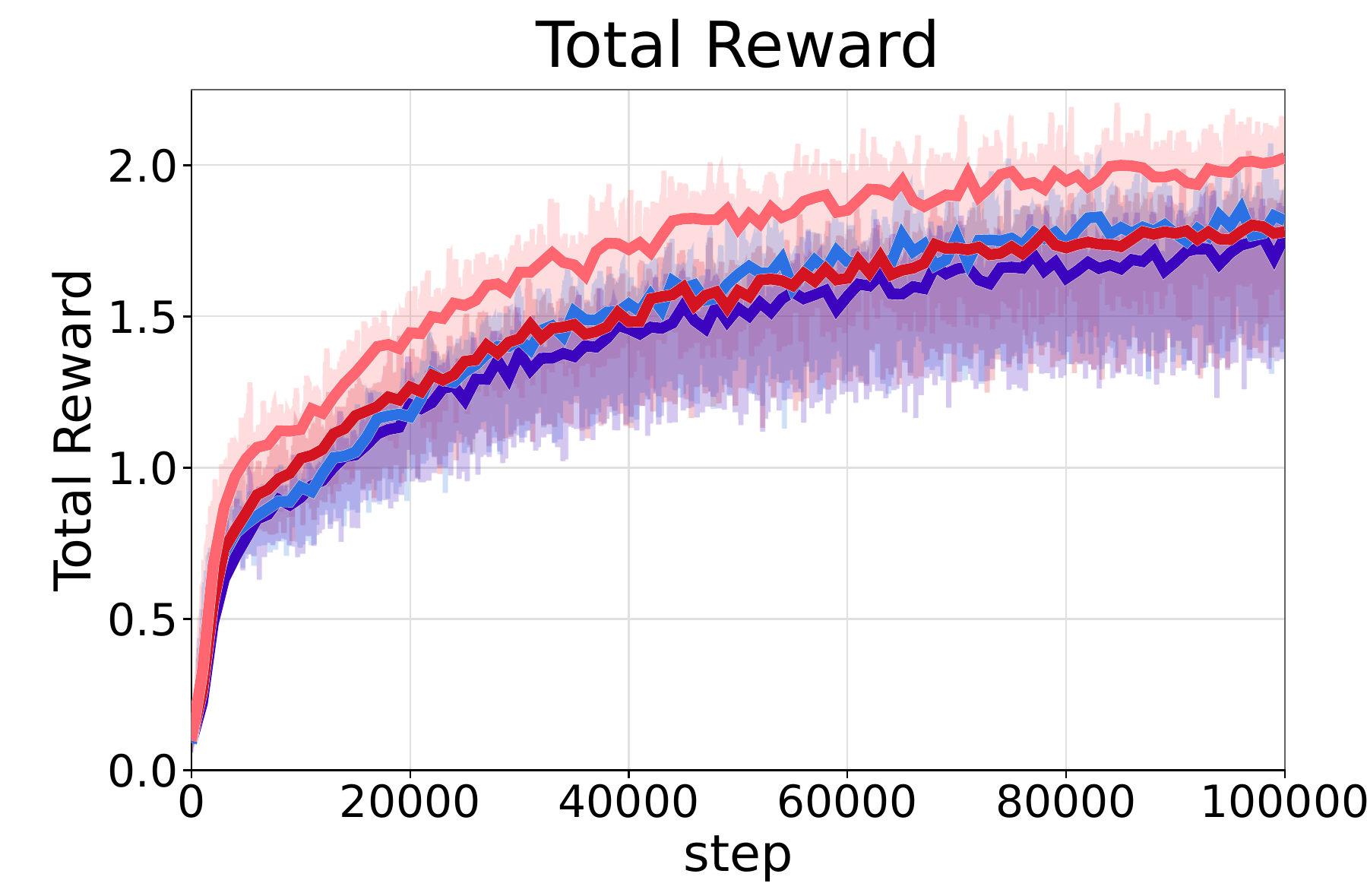}
\includegraphics[width=0.32\linewidth, trim= 0 0 0 0, clip]{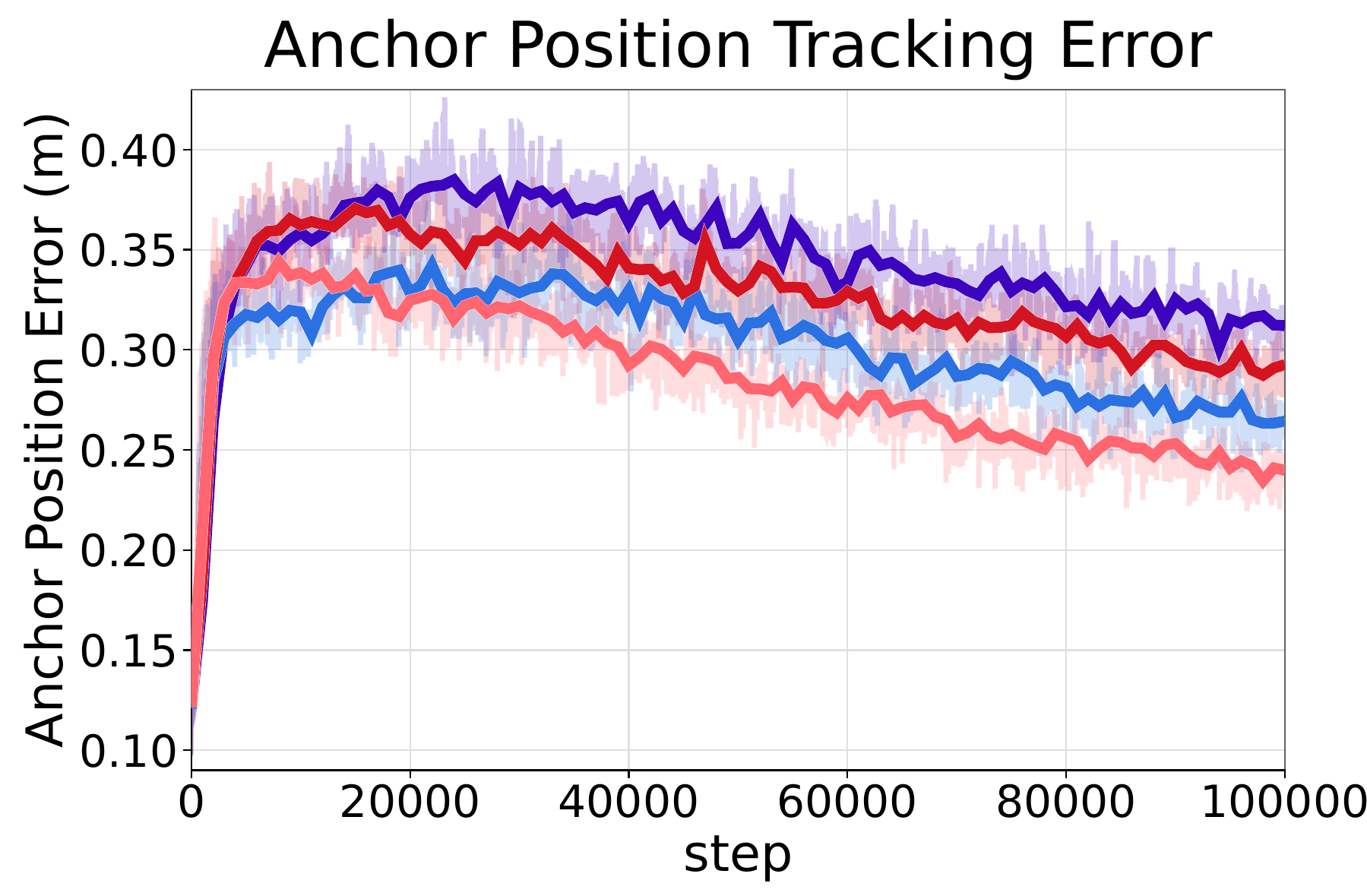}
\includegraphics[width=0.32\linewidth, trim= 0 0 0 0, clip]{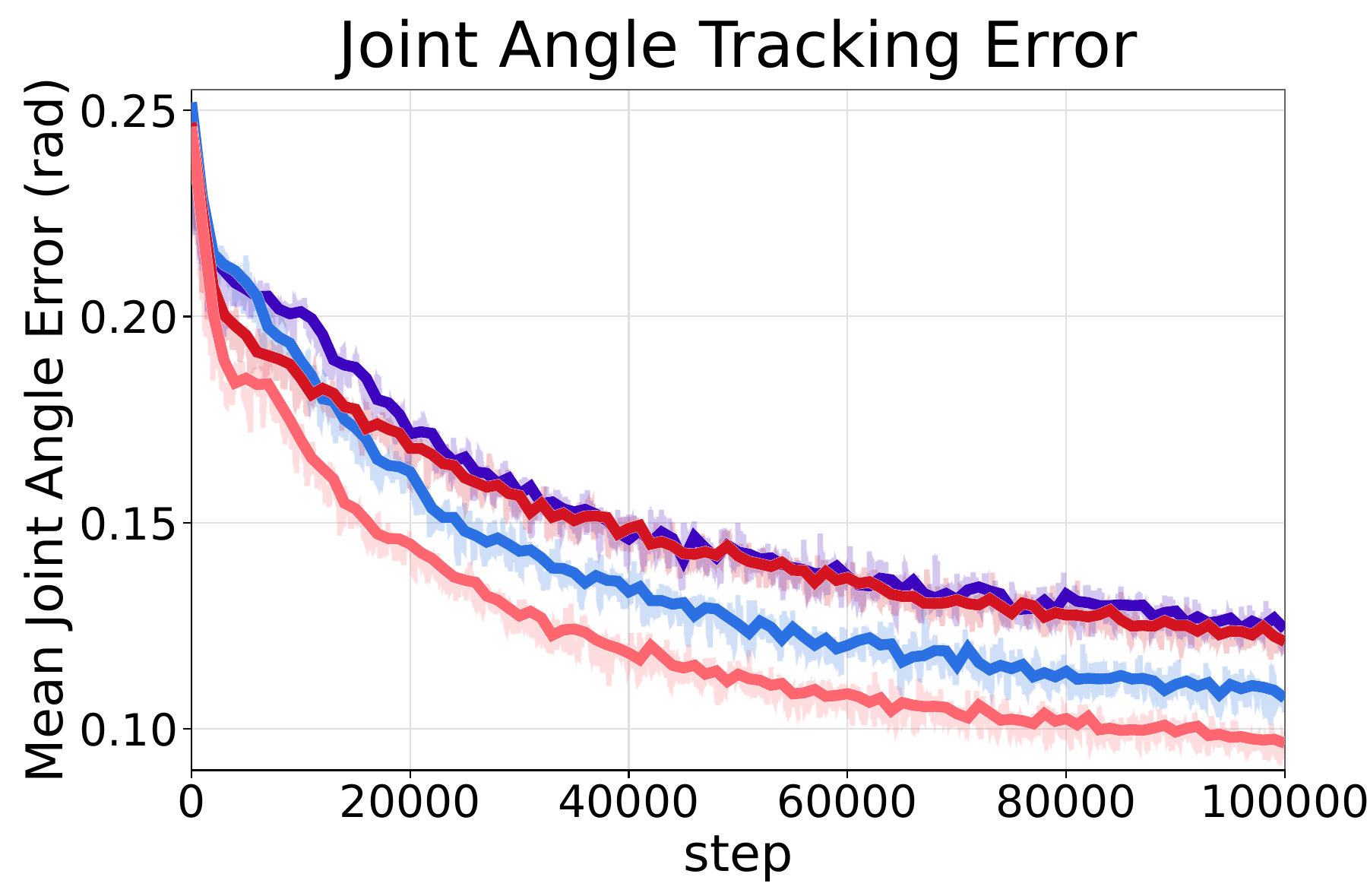}
\includegraphics[width=0.7\linewidth, trim= 0 5 0 4, clip]{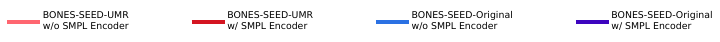}
\caption{SONIC training curves on BONES-SEED, comparing UMR-retargeted and original reference motions with and without the SMPL encoder.}
\label{fig:sonic-training}
\end{figure*}

\begin{figure}[tbp]
\centering
\includegraphics[width=0.95\linewidth, trim= 0 0 0 0, clip]{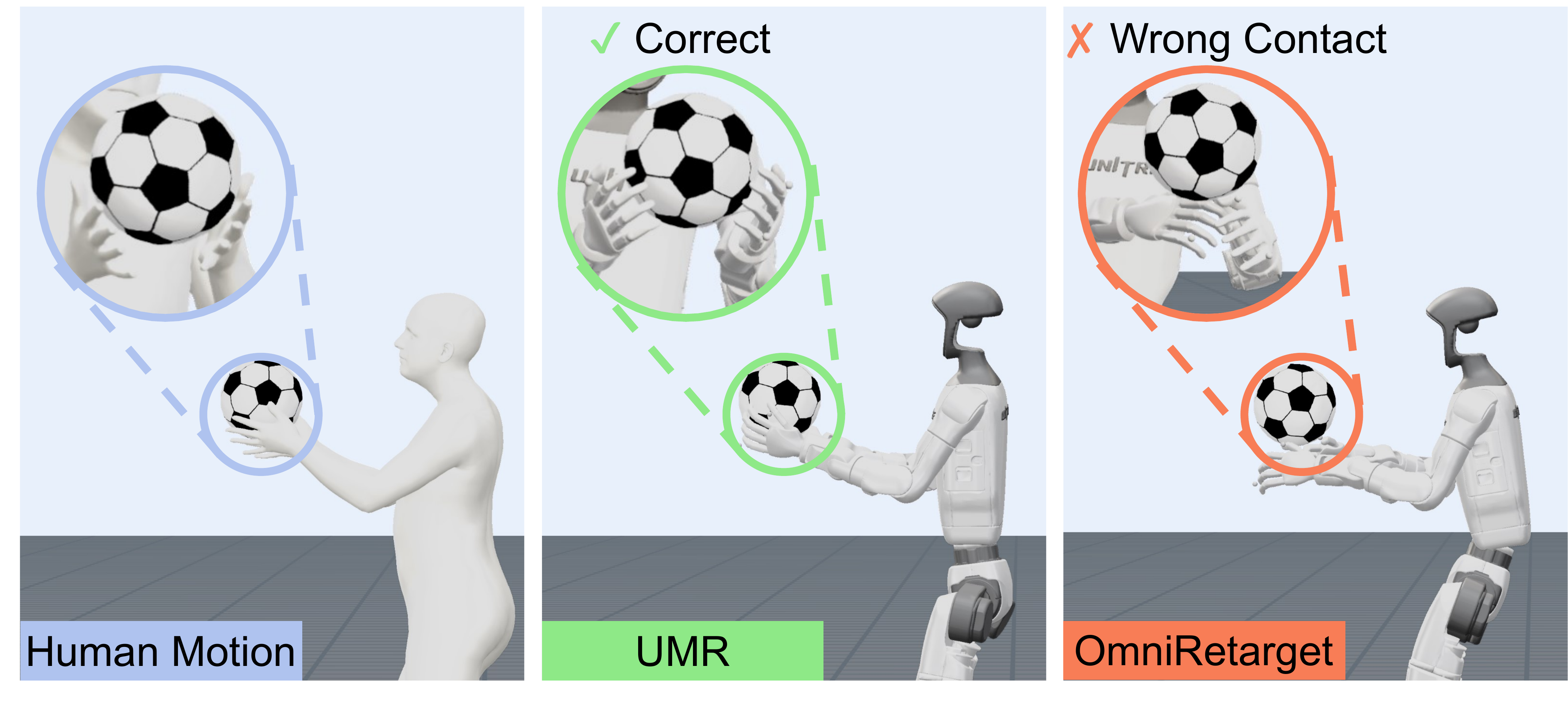}
\caption{\textbf{Interaction retargeting comparison.} For the same human reference (left), UMR preserves
  bimanual contact geometry (center), whereas OmniRetarget produces mismatched hand--object contacts (right).}
\label{fig:HOI-comparison}
\end{figure}

\textbf{Per-Motion Tracking Quality.}
Following GMR~\cite{gmr}, we evaluate the 40 LAFAN1 motion sequences for which Unitree retargeted references are available. We consider three evaluation settings: simulation without domain randomization (Sim w/o DR), simulation with domain randomization (Sim w/ DR), and Sim2Sim following the deployment setup of GMR. We report motion-tracking success rate and tracking errors. A rollout is successful if it completes the reference window without triggering termination. Tracking errors include global body-part position error ($E_{\mathrm{g\text{-}mpbpe}}$, mm), root-relative body-part position error ($E_{\mathrm{mpbpe}}$, mm), and joint-angle error ($E_{\mathrm{mpjpe}}$, $10^{-3}$ rad).
 
Table~\ref{tab:beyondmimic-success-rates} shows that UMR achieves higher tracking success rates than GMR, with clear gains on challenging motions such as Fall and GetUp and Fight even in nominal simulation. The advantage becomes broader under domain randomization and Sim2Sim, where UMR outperforms GMR across all motion categories while remaining competitive with the Unitree retargeted references. This suggests that UMR produces more plausible references, which lead to more robust policy tracking under domain randomization and Sim2Sim.

The tracking errors in Table~\ref{tab:beyondmimic-tracking-errors} further support this observation. UMR consistently improves tracking accuracy over GMR, remains comparable to the Unitree retargeted references in body-part position, and achieves the lowest mean joint-angle errors in both settings. Together with the higher success rates, these results indicate that UMR yields references that support more accurate and robust downstream tracking. Beyond these quantitative evaluations, Fig.~\ref{fig:2real}(a) demonstrates high-fidelity real-world tracking of a high-dynamic spin kick motion retargeted from MimicKit~\cite{MimicKitPeng2025}, further validating the practical effectiveness of UMR.

\textbf{Large-Scale Policy Learning.}
We further evaluate the utility of UMR-retargeted data for large-scale training of general motion trackers. SONIC trains a control policy over large motion collections and supports multiple control modalities. To enable human-motion control, SONIC employs an SMPL encoder that maps human motion into the same latent space as robot motion, with cross-modal consistency losses aligning the two representations. We evaluate the performance of UMR under SONIC both with and without the SMPL encoder: the former additionally leverages source human motion and latent-space alignment, whereas the latter more directly reflects the quality of the retargeted robot references.

The released Unitree G1 references in BONES-SEED~\cite{studio2026bones} are retargeted from SOMA-Uniform~\cite{saito2026soma} using GMR~\cite{gmr}, whereas UMR directly retargets actor-specific SOMA-Proportional motions. A key advantage of UMR is its ability to preserve actor-specific body geometry without requiring separate retargeting configurations for different actors. As illustrated in Fig.~\ref{fig:saito2026soma-comparison}, UMR better preserves the target posture and contact relationships, whereas GMR introduces lower-body pose distortion and ground contact artifacts.

Figure~\ref{fig:sonic-training} shows how these differences affect large-scale policy learning. We report total reward, anchor-position tracking error, and mean joint-angle tracking error. With the SMPL encoder enabled, UMR and the released Unitree references yield comparable performance, as source human motion and latent-space alignment provide additional guidance beyond the robot references. Without the SMPL encoder, UMR consistently outperforms the released references across all three metrics, with roughly 10\% relative improvement toward the end of training. These consistent gains show that the advantage of UMR remains evident at dataset scale.

\begin{table}[t]
\caption{Downstream policy performance on contact-rich tasks with references generated by UMR and OmniRetarget~\cite{yang2025omniretarget}. Each entry reports UMR / OmniRetarget. Better values are shown in \textbf{bold}.}
\label{tab:interaction-policy}
\centering
\scriptsize
\setlength{\tabcolsep}{2pt}
\setlength{\arrayrulewidth}{0.4pt}
\setlength{\heavyrulewidth}{0.6pt}
\setlength{\aboverulesep}{0.4ex}
\setlength{\belowrulesep}{0.4ex}
\renewcommand{\arraystretch}{1.08}
\begin{tabular}{@{}
>{\hspace*{4pt}\raggedright\arraybackslash}m{0.10\columnwidth}
*{3}{>{\centering\arraybackslash}m{\dimexpr0.30\columnwidth-2\tabcolsep\relax}}
@{}}
\toprule
& \textbf{Success Rate (\%) $\uparrow$}
& \textbf{Joint Error (rad) $\downarrow$}
& \textbf{Object Error (m) $\downarrow$} \\
\cmidrule(l{5pt}r{5pt}){2-2}\cmidrule(l{5pt}r{5pt}){3-3}\cmidrule(l{2.5pt}r{2.5pt}){4-4}
\textbf{Task} & UMR / OmniRetarget & UMR / OmniRetarget & UMR / OmniRetarget \\
\hline
\multicolumn{4}{l}{\hspace*{2pt}\textit{Robot-Object Interaction (Retargeting from the OmniContact Dataset)}} \\
\hline
Carry & \textbf{99.28}/82.89 & \textbf{0.570}/1.030 & \textbf{0.081}/0.252 \\
Kick & \textbf{99.98}/83.13 & \textbf{0.619}/1.396 & \textbf{0.038}/0.215 \\
Push & \textbf{99.99}/99.87 & \textbf{0.630}/1.043 & \textbf{0.048}/{0.049} \\
\hline
\multicolumn{4}{l}{\hspace*{2pt}\textit{Robot-Scene Interaction (Retargeting from the GRAIL Dataset)}} \\
\hline
Chair & {75.24}/\textbf{79.67} & 0.315/\textbf{0.308} & N/A/N/A \\
Stair & \textbf{43.53}/11.01 & \textbf{0.229}/0.327 & N/A/N/A \\
Slope & \textbf{58.32}/52.37 & \textbf{0.152}/0.179 & N/A/N/A \\

\bottomrule
\end{tabular}
\end{table}

\subsection{Contact-Rich Interaction Retargeting}

Compared with pure motion tracking, contact-rich interactions additionally require preserving the geometric relationships between the robot and its interaction target. We compare UMR with OmniRetarget~\cite{yang2025omniretarget} on robot-object and robot-scene interactions using motions from OmniContact~\cite{yu2026omnicontact} and GRAIL~\cite{xie2026grail}, respectively. The downstream policies follow the OmniContact and OmniRetarget training protocols, with only the retargeted references changed. We report task success rate and joint tracking error, together with object tracking error for HOI tasks.

Fig.~\ref{fig:HOI-comparison} illustrates a representative HOI example. Although both methods reproduce the approximate hand position, OmniRetarget can shift the contact to an incorrect surface region, whereas UMR preserves the intended hand-object contact geometry. This difference is reflected in Table~\ref{tab:interaction-policy}, where UMR outperforms OmniRetarget across all reported metrics on Carry, Kick, and Push. Notably, the relatively permissive termination criterion in OmniContact makes success rate less sensitive to tracking fidelity. Policies trained on UMR references reduce joint error by approximately 40\% to 56\% across all three tasks, while those trained on OmniRetarget references can still move the object despite substantial deviation from the reference pose. This indicates that UMR provides higher-quality references for accurate interaction tracking. Fig.~\ref{fig:2real}(b) further demonstrates successful real-world deployment of the resulting object-interaction policy.

For HSI, GRAIL motions provide a more challenging source than motion-capture data, as the human trajectories are reconstructed from VFM-generated videos through 4D human-motion estimation and optimization, making them inherently more susceptible to reconstruction errors~\cite{xie2026grail}. Despite this noisier source, UMR substantially outperforms OmniRetarget on Stair and performs better on both metrics for Slope, while OmniRetarget retains a slight advantage on Chair. One possible reason is that OmniRetarget's heuristic stance detection and foot-sticking hard constraint may become mismatched across motion distributions, causing unintended contacts to persist. UMR avoids such hand-crafted stance heuristics and remains effective across these scene-interaction motions. The stair-traversal policy is further deployed on the physical robot in Fig.~\ref{fig:2real}(c). For detailed visual comparisons, please refer to the supplementary video.

\section{Conclusion}

In this paper, we presented Unified Motion Retargeting (UMR), a surface-centric framework for retargeting heterogeneous human motion to humanoid robots without manually specified skeletal correspondences. UMR learns dense correspondence between source and robot surface point clouds in a canonical pose and reuses the paired points as geometric anchors for constrained motion optimization and contact transfer. This surface-based formulation provides a common interface across motion representations and robot embodiments while preserving fine-grained pose and interaction geometry. Experiments across diverse sources and embodiments demonstrate the generality of UMR. On Unitree G1, its references improve downstream tracking over GMR, remain competitive with curated Unitree references, and substantially outperform OmniRetarget on most contact-rich tasks. The resulting policies also transfer successfully to real robots in high-dynamic motions and diverse interaction scenarios. These results show that dense surface correspondence offers an effective and scalable interface for converting heterogeneous human motion into robot-ready references.

A current limitation is that UMR assumes access to mesh-based source geometry and a canonical source template. Future work will extend the framework to less structured motion observations, as well as more articulated embodiments such as dexterous hands, and multi-agent interactions.

\bibliographystyle{IEEEtran}
\bibliography{references}

\end{document}